\documentclass[letterpaper, 10 pt, conference]{ieeeconf}  % Comment this line out if you need a4paper

\IEEEoverridecommandlockouts                              % This command is only needed if 
\usepackage{times}
\usepackage{multicol}
\usepackage[bookmarks=true]{hyperref}
\usepackage{tikz}
\usepackage{xcolor}
\usepackage{hyperref}
\usepackage{amssymb}
\usepackage{amsmath, amsfonts}
\usepackage{graphicx}
\usepackage{siunitx}
\usepackage{standalone}
\usepackage{booktabs}
\usepackage[ruled,vlined]{algorithm2e}
\usepackage{mdframed}
\usepackage{fancyvrb}
\usepackage{soul}
\usepackage{dsfont,mathabx}
\usepackage[font=footnotesize, skip=5pt]{caption}
\newtheorem{problem}{Problem}

\newcommand{\textquote}[1]{\textquotedblleft #1\textquotedblright}
\newcommand{\etal}{\MakeLowercase{\textit{et al. }}}
\newcommand{\transp}[2]{\mathcal{A}_{\parallel_{#1}^{#2}}}

\newcommand{\tphsmm}{\mathbf{\Theta}}

\def\normal{\mathcal{N}}
\def\R{\mathbb{R}}
\renewcommand{\vec}[1]{\boldsymbol{#1}}
\newcommand{\trsp}{\mathsf{T}}

\title{\LARGE \bf
	Learning and Sequencing of Object-Centric Manipulation Skills\\ for Industrial Tasks$^{1}$
}

\author{Leonel Rozo\textsuperscript{$\ast$}\thanks{\textsuperscript{$\ast$} Equal contribution}, Meng Guo\textsuperscript{$\ast$}, Andras G. Kupcsik\textsuperscript{$\ast$}, Marco Todescato, Philipp Schillinger, \\ Markus Giftthaler, Matthias Ochs, Markus Spies, Nicolai Waniek, Patrick Kesper, Mathias B\"urger% <-this % stops a space
	\thanks{$^{1}$Bosch Center for Artificial Intelligence, Renningen, Germany
		{\tt\scriptsize <first name(s)>.<last name>@de.bosch.com}}
}

\begin{document}
		
	\maketitle
	\thispagestyle{empty}
	\pagestyle{empty}

	\begin{abstract}
		Enabling robots to quickly learn manipulation skills is an important, yet challenging problem.
		Such manipulation skills should be flexible, e.g., be able adapt to the current workspace configuration.
		Furthermore, to accomplish complex manipulation tasks, robots should be able to sequence several skills and adapt them to changing situations.
		In this work, we propose a rapid robot skill-sequencing algorithm, where the skills are encoded by object-centric hidden semi-Markov models.
	    The learned skill models can encode multimodal (temporal and spatial) trajectory distributions.
		This approach significantly reduces manual modeling efforts, while ensuring a high degree of flexibility and re-usability of learned skills.
		Given a task goal and a set of generic skills, our framework computes smooth transitions between skill instances.
		To compute the corresponding optimal end-effector trajectory in task space we rely on Riemannian optimal controller. 
		We demonstrate this approach on a 7~DoF robot arm for industrial assembly tasks.
	\end{abstract}

	%========================================
	%========================================

	\section{Introduction}\label{sec:intro}
	Deploying service robots in highly flexible manufacturing sites is promising, but also challenging~\cite{bedaf2017multi}.
	The challenges arise in different sub-fields of robotics, e.g., perception~\cite{elfes1989using}, motion planning~\cite{lavalle2006planning}, mapping and navigation~\cite{murray2000using}, or human-robot interaction~\cite{goodrich2008human}.
	In this work, we tackle two specific problems, namely, flexible motion skills generation and skills sequencing in the context of industrial tasks, with an emphasis on assembly settings.  
	First, it is impossible for robot manufacturers to pre-program \emph{all} robot capabilities (referred to as \emph{skills}) that end users may require.
	To avoid inquiring engineers whenever a new skill is needed, it is crucial to provide an easy and efficient method with which laymen can teach the robot new skills.
	Simply recording and replaying a demonstrated trajectory is often insufficient, because changes in the environment, e.g., varying robot and/or object poses, would render any attempt unsuccessful.
	In other words, the robot needs to recognize the intentions behind these demonstrations and thus generalize over unforeseen situations.
	
	Many learning-from-demonstration (LfD) frameworks have shown great improvements in this aspect.
	Compared to hard-coded alternatives, they embed extracted knowledge into probabilistic models.
	Examples are probabilistic movement primitives (ProMPs)~\cite{paraschos2013probabilistic}, Stable
	Estimators of Dynamical Systems (SEDS)~\cite{Khansari2011seds}, Task-Parameterized Gaussian Mixture Models (TP-GMMs)~\cite{calinon2016tutorial}, and more recently, Kernelized MPs (KMPs)~\cite{Huang2019KMPs} and Conditional Neural MPs (CNMPs)~\cite{Yunus2019CNMPs}.  
	However, most of these approaches train models specifically for each skill instantiation, with the exception of CNMPs, which is able to encode multiple modes of operation for the same skill.
	For most of the aforementioned models, ``grasp the object from the top'' and ``grasp the object from the side'' are usually treated as two different skills, and thus, different models are trained.
	This not only greatly decreases the teaching efficiency, but also significantly limits the \emph{reusability} of each skill.
 	Instead, we consider object-centric skills that represent robot end-effector motions relative to objects of interest for the task space.
 	To encode and generate full end-effector pose trajectories, we exploit Riemannian-manifolds theory to compute the necessary statistics and retrieve smooth control references.
 	Recent work in robot learning and control showed the efficiency and flexibility of the Riemannian formulation~\cite{Zeestraten17riemannian,Jaquier19CoRLa}.
	
	Additionally, several skills often need to be performed in sequence to accomplish tasks with increased complexity.
    In this work, we assume such a sequence is given by the user and our focus is on the adaptation of each skill to the sequence and to the varying configurations.
	The problem of finding the right sequence is commonly referred to as the task planning problem~\cite{ghallab2004automated}.
	Logic-based planning frameworks such as PDDL~\cite{fox2003pddl2} gained popularity due to close resemblance to human reasoning.
	However, manual definition of the planning model, such as pre-conditions and effects of  all skills, quickly becomes impractical due to the large variation of skills in different applications.	
	
	In this paper we propose a generic motion planning framework for sequencing manipulation skills that are learned from demonstration.
    First, to teach a \emph{general} skill, only few human demonstrations are needed with different object configurations.
	These demonstrations are used to train an object-centric model for each skill, which builds local models of the demonstrations from the perspective of different coordinate systems.
    Moreover, given a sequence of skills to execute, a complete model is constructed by cascading the local models with updated parameters.
  	Lastly, during execution, this model is used to compute the most-likely reference trajectory for the robot under various workspace configurations.
	Our method significantly reduces human modeling efforts, while ensuring high-degree flexibility and re-usability of learned skills.

	The main contribution of this paper is twofold:
	%(i)~we enable a robot to learn and reproduce more general skills by extending existing work on task-parameterized probabilistic models on Riemannian manifolds.
	(i)~we present a novel algorithm for the sequencing of several general skills to fulfill a given task;
	(ii)~and propose a skill-sequencing method that finds the most-likely reference trajectory given only the initial system state and the desired task goal. 
	%(iii)~Finally, we demonstrate how the skill execution can be adapted in real time due to changing environmental situations.
	Our framework builds on a Riemannian-manifold formulation to provide robust learning and optimal control, overcoming inaccuracy and stability issues that arise when using Euclidean approximations.

	The remainder of this paper is organized as follows.
	Section~\ref{sec:related} reviews the state of the art in motion primitives and skills sequencing. 
	Section~\ref{sec:preliminary} presents some preliminaries on TP-GMMs and Riemannian geometry, essential tools in our work.
	Section~\ref{sec:problem} details the considered problem.
	Section~\ref{sec:framework} contains the main contributions.
	Experimental results are presented in Section~\ref{sec:experiments}.
	Section~\ref{sec:conclusion} concludes the work.

	%========================================
	%========================================
	\section{Related Work}\label{sec:related}
	%learning by demonstration
	Learning by demonstration is an intuitive and natural way to transfer human skills to robots, which  recently gained much attention~\cite{Osa2018Imitation}.   
	% TPGMM
	Gaussian Mixture Models (GMMs) provide an elegant probabilistic representation of motion skills.
	For instance, the work by Niekum \etal\cite{niekum2015learning} shows how to use them to extract important features from only few human demonstrations of particular skills.
	Successful applications can be found in humanoids~\cite{Osa2018Imitation}, human-robot collaborative manipulation~\cite{Rozo15IROS} and robot motion planning~\cite{ye2017guided}.
	Furthermore, task-parameterized GMMs (TP-GMMs)~\cite{calinon2016tutorial} provide a powerful extension to GMMs by incorporating observations from the perspective of different frames of reference.
	This allows the robot to automatically adapt movements to new situations and has shown reliable performance in human-robot collaborative transportation~\cite{Rozo15IROS} and robot bimanual sweeping~\cite{silverio2015learning}.
	However, most of the above approaches focus on learning a specific TP-GMM model for each single skill, e.g., \textquote{holding the cube above the cup}~\cite{calinon2016tutorial} or \textquote{transporting the object}~\cite{Rozo15IROS}.
	In contrast, we propose to learn one TP-GMM model for a general class of skills (without explicitly distinguishing them) and choose the particular instantiation of each skill during run-time, depending on the given high-level plan and task goal.

	% sequencing of skills
	% LfD approaches
	Given a set of skills, the next problem is how to combine them for successful execution of complex manipulation tasks. 
	Researchers have mainly used either LfD~\cite{manschitz2013learning,manschitz2015Sequencing,pastor2012Associative} or reinforcement learning (RL)~\cite{grave2013Sequencing,Stulp2012PI2Seq} to master skills sequencing, most of them employing dynamic movement primitives (DMPs) as skill representation. 
	Manschitz \etal\cite{manschitz2013learning} learn a sequence graph of skills from kinesthetic demonstrations, where a classifier drives transitions between skills.
	The authors extend this approach for bimanual settings~\cite{manschitz2015Sequencing}, where the task is represented by a set of concurrent sequence graphs of motion primitives. 
	Entry and exit probabilities determine the transition between consecutive skills conditioned on the environment state.
	%The initial environment state determines the sequence to follow during execution. 
	Pastor \etal~\cite{pastor2012Associative} learn DMPs along with a distribution of sensory patterns.  
	Skills sequencing is achieved by choosing pairs of DMPs whose initial and final sensory patterns closely match. 
	Our work distinguishes in that our skill representation encodes the diverse effects of robot actions on objects, which may differ according to the particular instantiation of the skill.
	Similarly to~\cite{pastor2012Associative}, our method exploits the distribution of observations to build a task model by cascading different skills as a function of their similarity in a Kullback-Leibler sense.  
	%The approach was tested in a reaching for a drill task with grasping actions to switch it on.
	
	%% RL-based approaches
	Gr\"{a}ve and Behnke~\cite{grave2013Sequencing} combine LfD and RL to learn a sequence of skills in an active learning setting. 
	%% An action-value function approximated by a Gaussian process provides information about unknown states and unsafe actions through the model uncertainty. 
	%% When the robot is uncertain, a human demonstrates a full sequence of actions to fulfill the task. 
	%% Otherwise, the robot improves its policy autonomously.
        %% %Tested in simulated pick-and-place task for a 3-skills sequencing setting. 
	In~\cite{Stulp2012PI2Seq}, a modified PI$^2$ method adapts the shape and attractor of several learned DMPs to smoothly sequence them.  
	%% PI$^2$ uses a cost-to-go function that considers both the trajectories of the current DMP and the subsequent ones.
	A similar approach is proposed in~\cite{lioutikov2016learning} to sequence DMPs in bimanual manipulation.  
	%Pick and place task with 2 DMPs was used to test the approach, where the grasping point for the manipulated object varied according to the final location where it was supposed to be placed.   
	In~\cite{konidaris2012robot}, the sequencing problem is investigated from a hierarchical RL perspective, where each skill is represented by a general control policy. 
	The work by M\"{u}lling \etal~\cite{muelling2010learning} proposes a gating network that activates the appropriate skill among a set of learned DMPs for table tennis.
	Lastly, Kroemer \etal~\cite{kroemer2015towards} use model-based RL to learn a high-level policy that sequences learned motion skills. 
	Most of the above approaches require human demonstrations for the complete task.
	Moreover, RL is used to adapt the skill parameters to make skills sequencing possible. 
	In contrast, our approach requires only demonstrations on the level of \emph{individual} skills. 
	RL adaptation of skill parameters for sequencing is unnecessary as our method builds a complete model of the task that considers the different skill instantiations and exploits predictive models of the skill effects.

	%========================================
	%========================================
	\section{Preliminaries}\label{sec:preliminary}
	We briefly present some preliminary results in robot skill learning, in particular, TP-GMMs, hidden semi-Markov models (HSMMs), and Riemannian manifolds.

	%========================================
	\subsection{TP-GMMs}\label{subsec:tp-gmm}
	%% The basic idea of ``learning from demonstrations'' is to fit a prescribed skill model such as GMMs to a handful of demonstrations using e.g., Expectation Maximization (EM). 
	%% %collect a handful of demonstrations from the manipulator perspective and fit, e.g., using Expectation Maximization (EM), a prescribed model, say a Gaussian Mixture Model (GMM) with a predefined number of components. 
	%% \emph{Task-parametrized} methods extend these standard models to to embed task-related parameters.
	%% %As an example, during a picking task, the position of the object itself w.r.t., e.g., the base frame of the manipulator as well as the point where the object is grasped w.r.t., e.g., its reference frame, can represent desirable task parameters. 
	%% Task parameters are closely related to the objects to interact with. 
	%% %The hope is that by learning a model parametrized w.r.t. object location and grasping point will lead (with some additional computations) to a generalized skills which can automatically adapt to a variety of tasks, when grounded/queried on different locations and grasping points.
	%% The introduction of task parameters greatly improves the adaptability of the learned skills. 
	%% In the rest of the paper, we adopt a specific class of models, namely \emph{Task Parametrize Gaussian Mixture Models} (TP-GMMs)  
	%% %In what follows, we briefly summarize the core aspects.
	The basic idea of LfD is to fit a prescribed skill model such as GMMs to a handful of demonstrations. 
	We assume we are given $M$ demonstrations, each of which contains $T_m$ data points for a dataset of $N=\sum_m T_m$ total observations $\vec\xi=\left\{\vec \xi_t\right\}_{t=1}^{N}$, where~$\vec\xi_t\in\R^d$.
	Also, we assume the same demonstrations are recorded from the perspective of $P$ different coordinate systems (given by the task parameters, such as, objects of interest).
	One common way to obtain such data is to transform the demonstrations from a static global frame to frame~$p$ by $\vec \xi_t^{(p)} = \vec A^{(p)^{-1}}(\vec \xi_t - \vec b^{(p)})$. 
	Here, $\{(\vec b^{(p)},\vec A^{(p)})\}_{p=1}^P$ is the translation and rotation of frame $p$ w.r.t. the world frame. %which we assume is static across the demonstration.
	Then, a TP-GMM is described by the model parameters $\{\pi_k,\{\vec\mu_k^{(p)},\vec\Sigma_k^{(p)}\}_{p=1}^P\}_{k=1}^K$  
	%\begin{equation}\label{eq:tp-gmm}
	%\left\{\pi_k,\{\vec\mu_k^{(p)},\vec\Sigma_k^{(p)}\}_{p=1}^P\right\}_{k=1}^K \,,
	%\end{equation}
	where $K$ represents the number of Gaussian components in the mixture model, $\pi_k$ is the prior probability of each component, and $\{\vec\mu_k^{(p)},\vec\Sigma_k^{(p)}\}_{p=1}^P$ are the parameters of the $k$-th Gaussian component within frame~$p$. 
	Differently from standard GMM, the mixture model above can not be learned independently for each frame. 
	Indeed, the mixing coefficients $\pi_k$ are shared by all frames and the \mbox{$k$-th} component in frame $p$ must map to the corresponding \mbox{$k$-th} component in the global frame. 
	Expectation-Maximization (EM)~\cite{calinon2016tutorial} is a well-established method to learn such models.
	Task parameterization of GMMs incorporates observations from the perspective of different frames of reference, thus allowing the robot to automatically adapt its motion to new situations.
	
	%% At this point, similar to standard GMMs learning described in~\cite{niekum2015learning}, it is possible to learn a mixture model for each frame.
	%% In particular, assume that a mixture model with $K$ components is to be learned in the world frame, each of which has a prior probability $\pi_k$. 
	%% Then, for frame $p$ this corresponds to a model of the type
	%% $p(\vec \xi^{(p)}) = \sum_{k=1}^K \pi_k\ \normal(\vec \xi^{(p)} |\vec  \mu_k^{(p)}, \vec \Sigma_k^{(p)})$, where $$
	
	%% leading, for all the coordinate systems, to a TP-GMM characterized by $\left\{\pi_k,\left\{\vec\mu_k^{(p)},\vec\Sigma_k^{(p)}\right\}_{p=1}^P\right\}_{k=1}^K$ parameters which can be, e.g., learned jointly
	%% \footnote{During learning it is important to take care of two aspects. 
	%% First when training the parameters of a certain coordinate systems only demonstrations recorded in that frame should be used.
	%% Second, the mixing coefficients must be shared by all the different GMMs.} 
	%% by ML using EM.

	Once learned, the TP-GMM can be used during execution to reproduce a trajectory for the learned skill.
	Namely, given the observed frames $\{\vec b^{(p)},\vec A^{(p)}\}_{p=1}^P$, the learned TP-GMM is converted into one single GMM with parameters $\{\pi_k, (\vec{\hat{\mu}}_{k}, \vec{\hat{\Sigma}}_{k})\}_{k=1}^K$, by multiplying the affine-transformed Gaussian components across different frames,
	%\footnote{Combining the mixture models expressed in different coordinate systems by multiplying the corresponding components is one straightforward way to do that and not the only possibility.}
	%Namely, it is possible to compute
	%
	%\[
	%p(\vec{\hat{\xi}}_t) = 
	%\sum_{k=1}^K \pi_k\ \normal(\vec{\hat{\mu}}_{t,k}, \vec{\hat{\Sigma}}_{t,k})
	%\]
	as follows 
	\begin{equation}\label{eq:gaussian-product}
	%\begin{split}
	%&    \left(\vec{\hat{\Sigma}}_{t,k}\right)^{-1} = \sum_{p=1}^P\left(\vec{\hat{\Sigma}}_{t,k}^{(p)}\right)^{-1}, \\
	%&    \vec{\hat{\mu}}_{t,k} = \vec{\hat{\Sigma}}_{t,k}\sum_{p=1}^P\left(\vec{\hat{\Sigma}}_{t,k}^{(p)}\right)^{-1}\vec{\hat{\mu}}_{t,k}^{(p)} \,,
	%\end{split}
	\vec{\hat{\Sigma}}_{k}=\left[ \sum_{p=1}^P\left(\vec{\hat{\Sigma}}_{k}^{(p)}\right)^{-1}\right]^{-1}, \vec{\hat{\mu}}_{k}=\vec{\hat{\Sigma}}_{k}\left[\sum_{p=1}^P\left(\vec{\hat{\Sigma}}_{k}^{(p)}\right)^{-1}\vec{\hat{\mu}}_{k}^{(p)}\right],
	\end{equation}
	where the parameters of the updated Gaussian at each frame $p$ are computed as $\vec{\hat{\mu}}_{k}^{(p)} = \vec A^{(p)} \vec\mu_k^{(p)} + \vec b^{(p)}$ and ${\vec{\hat{\Sigma}}_{k}^{(p)} = \vec A^{(p)} \vec\Sigma_k^{(p)} \vec A^{(p)^\trsp}}$. 
	While the task parameters may vary over time, we dropped the time index for the sake of notation.
	\subsection{HSMMs}\label{subsec:hsmm}
	Hidden semi-Markov Models (HSMMs) extend standard hidden Markov Models (HMMs) by embedding temporal information of the underlying stochastic process. 
	That is, while in HMM the %underlying hidden process is assumed to be Markov, i.e., the 
	probability of transitioning to the next state depends only on the current state, in HSMM %the state process is assumed semi-Markov.
	%This means that a 
	this transition also depends on the elapsed time since the state was entered.
%	HSMMs have been extensively applied in the past in speech synthesis, see \cite{ZEN2007}. 
	HSMMs have been successfully applied, in combination with TP-GMMs, for robot skill encoding to learn spatio-temporal features of the demonstrations~\cite{Tanwani2016}.
	More specifically, 
	%An HSMM model explicitly encodes state duration probabilities in Gaussian distributions, which can be jointly learned with state models using EM. 
	a task-parameterized HSMM (TP-HSMM) model is defined as: 
	\begin{equation*}
	\tphsmm = \left\{ \{a_{hk}\}_{h=1}^K, (\mu_k^D, \sigma_k^D), \pi_k, \{(\vec \mu_k^{(p)}, \vec\Sigma_k^{(p)})\}_{p=1}^P \right\}_{k=1}^K\,,
	\end{equation*}
	where $a_{hk}$ is the transition probability from state~$h$ to $k$; $(\mu_k^D, \sigma_k^D)$ describe the Gaussian distributions for the duration of state $k$, i.e., the probability of staying in state $k$ for a certain number of consecutive steps; $\{\pi_k, \{\vec \mu_k^{(p)}, \vec \Sigma_k^{(p)}\}_{p=1}^P\}_{k=1}^K$ equal the TP-GMM introduced earlier, representing the observation probability corresponding to state $k$. Note that in our HSMM the number of states corresponds to the number of Gaussian components in the \textquote{attached} TP-GMM.
	
	Given a certain (partial) sequence of observed data points $\{\vec\xi_\ell\}_{\ell=1}^t$, assume that the associated sequence of states in~$\tphsmm$ is given by $\vec s_t = s_1s_2\cdots s_t$.
	As shown in \cite{Tanwani2016}, the probability of data point $\vec\xi_t$ belonging to state $k$ (i.e., $s_t=k$) is given by the \emph{forward} variable ${\alpha_t(k)=p(s_t = k, \{\vec\xi_\ell\}_{\ell=1}^t)}$: 
	\begin{equation}\label{eq:forward_variable}
	\alpha_t(k) = \sum_{\tau=1}^{t-1} \sum_{h=1}^K \alpha_{t-\tau}(h) a_{hk}\, \normal(\tau|\mu_k^D, \sigma_k^D)\, o_{\tau}^t\,,
	\end{equation}
	where $o_{\tau}^t = \prod_{\ell=t-\tau+1}^t \normal(\vec \xi_{\ell}|\hat{\vec \mu}_{k},\hat{\vec \Sigma}_{k})$ is the emission probability and $(\hat{\vec \mu}_{k},\hat{\vec \Sigma}_{k})$ are derived from~\eqref{eq:gaussian-product} given the task parameters.
	%which can be interpreted as
	%
	%\begin{equation*}
	%\begin{split}
	%\alpha_t(i) &= \sum_{j=1}^K a_{ji} \sum_{\tau=1}^{t-1}p(\vec \xi_1,\ldots,\vec \xi_{t-\tau}, s_{t-\tau}=j) \\
	%&\qquad p(\tau|s=i) p(\vec \xi_{t-\tau+1},\ldots,\vec \xi_t|s=i)
	%\end{split}
	%\end{equation*}
	%
	%meaning the probability of transitioning from any other state $j$ into $i$ and then remanining in state $i$ for $\tau$ time steps. 
	%Using the forward variable is then possible to compute the most probable sequence of state as
	%
	%\[
	%\{s_1,\ldots,s_T\} = \arg\max_i\ \alpha_t(i)
	%\]
	Furthermore, the same forward variable can also be used during reproduction to predict future steps until $T_m$.

	In this case however, since future observations are not available, only transition and duration information are used as explained by~\cite{yu2003missingHSMM}, i.e., by setting $\normal(\vec \xi_\ell|\hat{\vec \mu}_{k},\hat{\vec \Sigma}_{k}) = 1$ for all $k$ and $\ell>t$ in~\eqref{eq:forward_variable}.
	%% \[
	%% \alpha_t(i) = \sum_{\tau=1}^{t-1} \sum_{j=1}^K \alpha_{t-\tau}(j) a_{ji} \normal(\tau|\mu_i^D, \sigma_i^D).
	%% \]
	At last, the sequence of the most-likely states $\vec s^\star_{T_m}=s_1^\star s_2^\star\cdots s_{T_m}^\star$ is determined by choosing $s^\star_t=\arg \max_k \alpha_t(k)$, $\forall 1\leq t\leq T_m$.	
	
	%% After retrieving the optimal state sequence $\{s_1,\dots,s_T\}$, one way to reproduce the motion is by minimum intervention control. 
	%% That is, we aim to minimize the control objective
	
	%% \begin{eqnarray*}
	%%  c(\vec u) &=& \sum_{t=1}^T (\vec \xi_t - \vec{\hat{\mu}}_{s_t})^T \vec{\hat{\Sigma}}_{s_t}^{-1}(\vec \xi_t - \vec{\hat{\mu}}_{s_t}) + \vec u_t^T \vec R \vec u_t,\\
	%%  && \mbox{s.t. } \vec \xi_{t+1} = \vec A \vec \xi_t + \vec B \vec u_t,
	%% \end{eqnarray*}
	
	%% where $\vec A$ and $\vec B$ are system dynamics parameters and $\vec u$ is the control effort. 
	%% To retrieve only the reference trajectory we can assume double integrator dynamics, which provides smooth state transitions.

	%========================================
	\subsection{Riemannian Manifolds}\label{sec:riemannian}
	
	\begin{figure}[t]
		\centering
		\includegraphics[width=.45\linewidth]{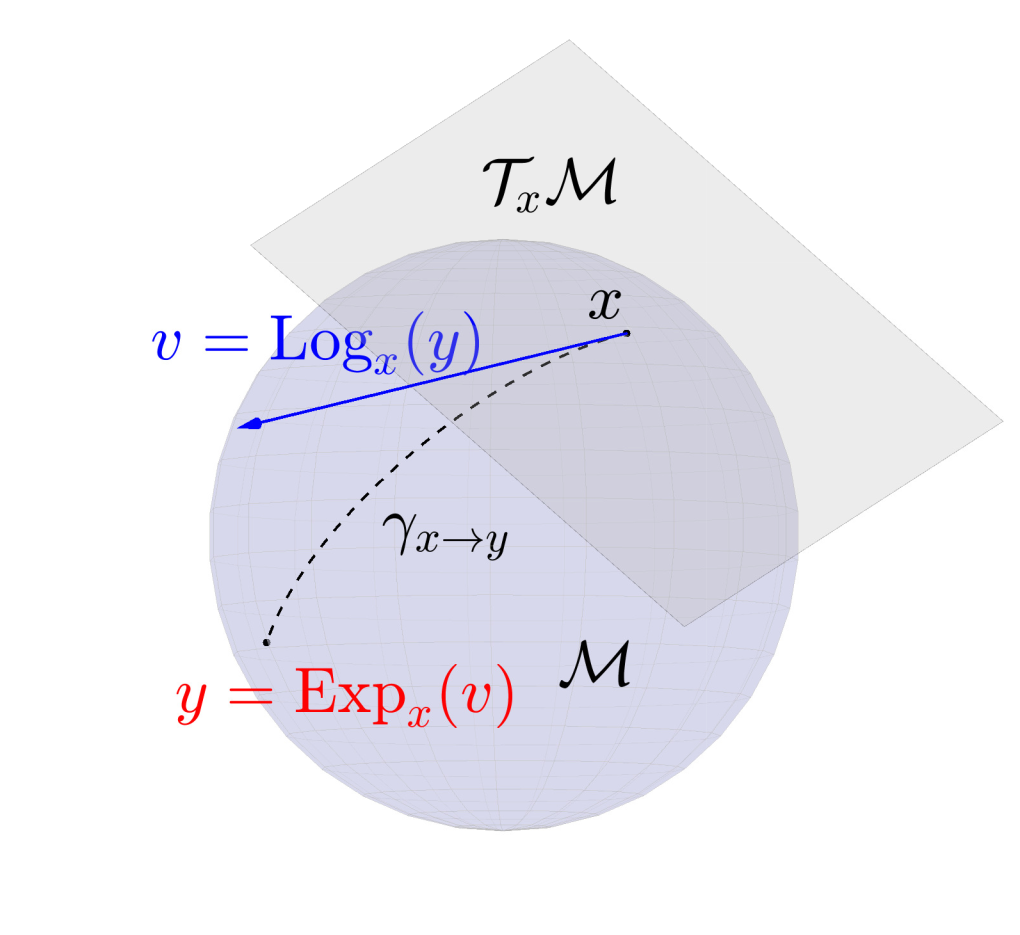}
		\includegraphics[width=.45\linewidth]{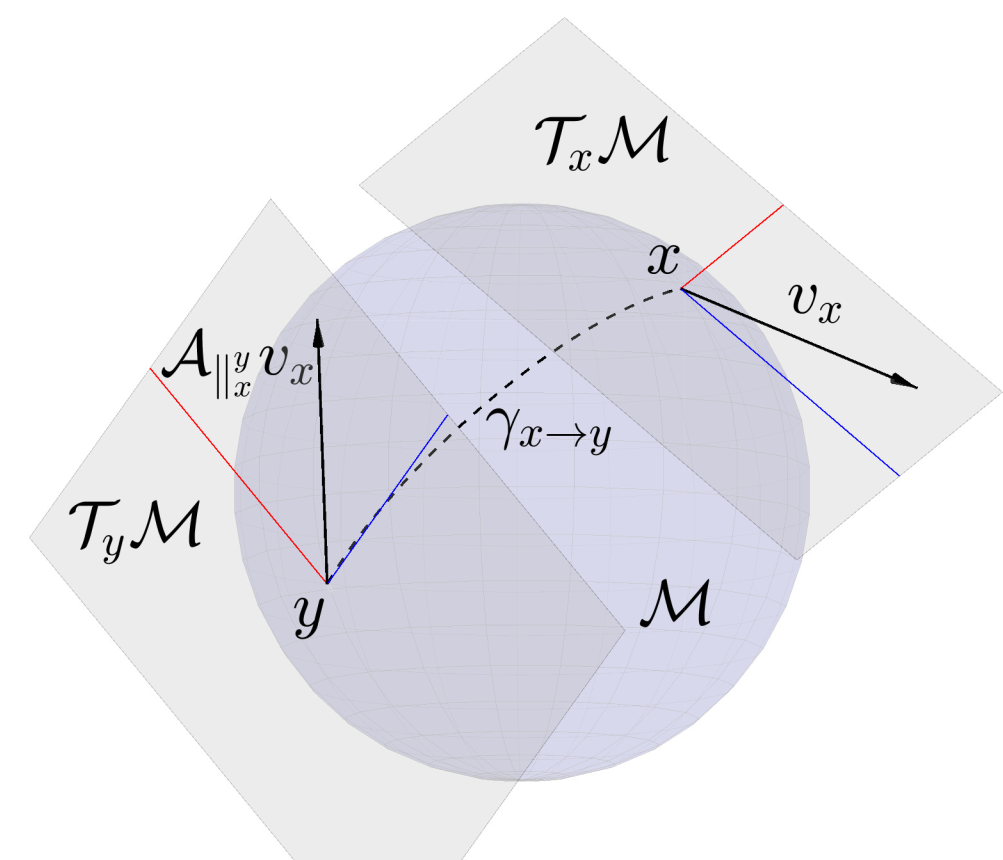}		
		\caption{\textbf{Left:} the illustration of the Log and Exp maps with geodesic $\gamma_{x \rightarrow y}$. Note that $\left\| v \right\|_2 = \left\| \gamma_{x \rightarrow y} \right\|_2$. \textbf{Right:} illustration of the parallel transport operation. $v_x$ is a vector defined in the tangent space of $x$, while the parallel transported vector $\mathcal{A}_{\parallel_x^y}v_x$ will lie in the tangent space of $y$ and is considered parallel to $v_x$.}
		\label{fig:riemannian}
		\vspace{-0.4cm}
	\end{figure}
	
	As the robot motion skills are learned from and reproduce time-varying poses of the end-effector, classical Euclidean-based methods are inadequate for processing such data, as they rely on rough approximations to account for the constraints imposed by orientation representation such as quaternions.
	These approximations may lead to inaccurate skill models or unstable controllers. 
	We instead endow the robot task space with a Riemannian manifold $\mathcal{M}$~\cite{Zeestraten17riemannian}.
	Briefly, for each point $\vec x$ in the manifold $\mathcal{M}$, there exists a tangent space $\mathcal{T}_{\vec x} \mathcal{M}$.
	This allows us to carry out Euclidean operations locally, while being geometrically consistent with manifold constraints. 
	
	We use exponential and logarithmic maps to map points between $\mathcal{T}_{\vec x} \mathcal{M}$ and $\mathcal{M}$ (Fig. \ref{fig:riemannian} left).
	The exponential map ${\text{Exp}_{\vec x}: \mathcal{T}_{\vec x} \mathcal{M}\to \mathcal{M}}$ maps a point in the tangent space of point $\vec{x}$ to a point on the manifold, while maintaining the geodesic distance.
	The inverse operation is called the logarithmic map $\text{Log}_{\vec x}: \mathcal{M}\to \mathcal{T}_{\vec x}\mathcal{M}$.
	Another useful operation is the parallel transport $\transp{\vec x}{\vec y}: \mathcal{T}_{\vec x}\mathcal{M}\to\mathcal{T}_{\vec{y}}\mathcal{M}$, which moves elements between tangent spaces without introducing distortion (Fig. \ref{fig:riemannian} right).
	The exact form of the aforementioned operations depend on the Riemannian metric associated to the manifold, which in our case corresponds to the formulations in~\cite{Zeestraten17riemannian}. 

	In this paper we exploit Riemannian manifolds to: \textbf{(a)}~properly compute statistics over $\mathcal{M}$ using Riemannian normal distributions that encode full end-effector motion patterns~\cite{Zeestraten17riemannian}; and \textbf{(b)} retrieve a smooth reference trajectory corresponding to the task plan (i.e., sequenced skills) using Riemannian optimal control, as proposed in Section~\ref{subsubsec:reproduce-traj-single}. 
	
	%========================================
	\begin{figure*}[h]
		\centering
		\includegraphics[width=0.84\textwidth]{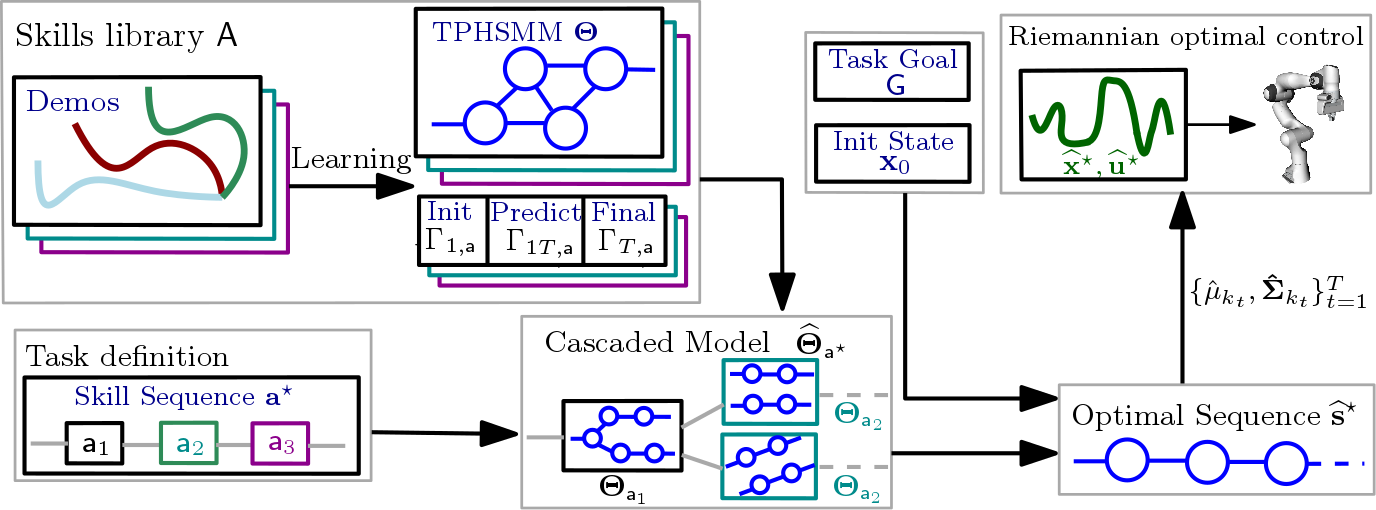}
		\caption{Overall diagram of the proposed method. From multiple kinesthetic demonstrations we encode object-centric skill models into TP-HSMMs and learn precondition/prediction/effect models $\boldsymbol{\Gamma}_{\mathsf{a}}$. Given a high-level task definition, we cascade the learned skill models into a joint model $\widehat{\mathbf{\Theta}}_{\mathsf{a}^\star}$. Based on the desired goal $\mathsf{G}$ of the task and the initial state of the environment, we find the optimal state-sequence $\widehat{\mathbf{s}}^\star$ using a modified Viterbi algorithm. Finally, to generate the corresponding reference trajectory we rely on a Riemannian optimal controller.}
		\label{fig:diagram}
		\vspace{-0.4cm}
	\end{figure*}
	%========================================
	
%==============================
	\section{Problem Description}\label{sec:problem}
	Consider a multi-DoF robotic arm, whose end-effector has state $\vec{x}_{\mathsf{e}} \in \mathbb{R}^3 \times \mathcal{S}^3 \times \mathbb{R}^1$ (describing the Cartesian position, orientation quaternion and gripper state), that operates within a static and known workspace.
	Also, within the reach of the arm, there are objects of interest denoted by ${\mathsf{O}=\{\mathsf{o}_1, \mathsf{o}_2,\cdots,\mathsf{o}_J\}}$, each of which has state ${\vec{x}_{\mathsf{o}_j}\in \mathbb{R}^3 \times \mathcal{S}^3}$.
	For simplicity, the overall system state is denoted by $\vec{x}=\{{\vec{x}_{\mathsf{e}}}, \{\vec{x}_{\mathsf{o}_j}, \forall \mathsf{o}_j\in \mathsf{O}\}\}$.
	
	Within this setup, an operator performs several kinesthetic demonstrations on the arm to manipulate one or several objects for certain manipulation skills.
	Denote by ${\mathsf{A}=\{\mathsf{a}_1,\mathsf{a}_2,\cdots,\mathsf{a}_H\}}$ the set of demonstrated skills. 
	Moreover, for skill $\mathsf{a}\in \mathsf{A}$, the set of objects involved is given by $\mathsf{O}_{\mathsf{a}}$ and the set of available demonstrations is denoted by~$\mathsf{D}_{\mathsf{a}}$.
	Note that all demonstrations follow the object-centric structure introduced in Section~\ref{subsec:tp-gmm}, i.e., they are recorded from multiple frames, often associated to the pose of the objects in $\mathsf{O}_{\mathsf{a}}$.
	For example, %the skill \textquote{pick up a peg} involves the object \textquote{peg} and the associated demonstrations are recorded from both the robot frame and the \textquote{peg} frame; another 
	the skill \textquote{insert the peg in the cylinder} involves the objects \textquote{peg} and \textquote{cylinder},         and the associated demonstrations are recorded from the robot, the \textquote{peg} and the \textquote{cylinder} frames.
    Note that one general skill can include several execution instances, e.g., ``pick the peg'' includes ``pick the peg from top, left or right'', and ``drop the peg'' comprises ``drop the peg inwards or outwards''.
    
    The manipulation tasks we consider consist of a \emph{given} sequence of skills $\vec{\mathsf{a}}^\star$ chosen from the demonstrated skills $\mathsf{A}$.
    For example, an insertion task involves ``pick the cap, re-orient the cap, pick the cap again and the insert the cap''.
    In the end of the task, a goal configuration $\mathsf{G}$ is reached as the desired final state of the system, including the robot and the objects.
    The considered problem is as follows:
	\begin{problem}\label{prob:overall-statement}
		Given a set of demonstrated skills $\mathsf{A}$, the desired skill sequence $\vec{\mathsf{a}}^\star$ and the goal $\mathsf{G}$, the objective is twofold: 
		\textbf{(a)} to reproduce each skill in $\vec{\mathsf{a}}^\star$ for a given goal state;
		and \textbf{(b)} to subsequently reproduce the sequence~$\vec{\mathsf{a}}^\star$ to maximize the success rate of achieving the goal $\mathsf{G}$. \hfill $\blacksquare$
	\end{problem}
%	Note that, in this work, motion planning is performed at the end-effector/gripper trajectory level. 
%	This means it is assumed that a low-level motion controller is used to track the desired trajectory; see~\cite{sciavicco2012modelling} for more details.

	%========================================
	%========================================
	\section{Learning and Sequencing of Flexible Skills}\label{sec:framework}

	In this section, we present the  main building blocks of the proposed framework, regarding the two objectives above. 
	In Figure \ref{fig:diagram} we show the diagram of our proposed approach.

	%========================================
	\subsection{Single Skill Reproduction} \label{subsec:skill-reproduce}
	Consider one demonstrated skill $\mathsf{a} \in \mathsf{A}$, associated with the set of demonstrations  $\mathsf{D}_{\mathsf{a}}=\left\{\vec \xi_t\right\}_{t=1}^{N}$, recorded in $P$ frames.
    Note that such frames are directly attached to the objects in $\mathsf{O}_{\mathsf{a}}$.
    As mentioned in Section~\ref{subsec:hsmm}, given a properly chosen number of components $K$, the TP-HSMM model $\tphsmm_{\mathsf{a}}$ abstracting the spatio-temporal features of trajectories related to skill $\mathsf{a}$, can be learned using an EM-like algorithm.
	It is worth emphasizing that, conversely to most existing work that only treats each instance of the same skill separately, we construct one model for the general skill. 
	To give an example, Fig.~\ref{fig:hsmm} shows an illustration of an HSMM for ``pick the peg'' skill that contains $10$ demonstrations for ``pick from top'' and ``pick from side''. 
	The learned HSMM in the global frame has a single initial state from which two branches encode the two different instances of the same ``pick'' skill.

	%%%%%%%%%%%%%%%%%%%%%%%%%%%%%%%
	\subsubsection{\textbf{Optimal state sequence}}\label{subsubsec:optimal-sequence-single}
	Consider now that a desired final observation of the robot state is given as $\vec \xi_{T}$, where $T$ is the skill time horizon (e.g. the average length over the demonstrations).
        Moreover, the initial robot state is observed as $\vec \xi_{1}$.
        We firstly address the following sub-problem:
	\begin{problem}\label{prob:reproduce-given-final}
	  Given the learned model $\tphsmm_{\mathsf{a}}$, construct the most-likely state sequence $\vec s^\star_T$ given \emph{only} $\vec \xi_1$ and $\vec \xi_{T}$. \hfill $\blacksquare$
	\end{problem}

\begin{figure}[t]
		\centering
		\includegraphics[height=0.35\linewidth]{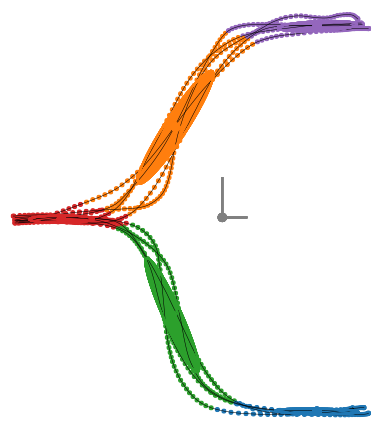}
		\hspace{0.2in}
		\includegraphics[height=0.35\linewidth]{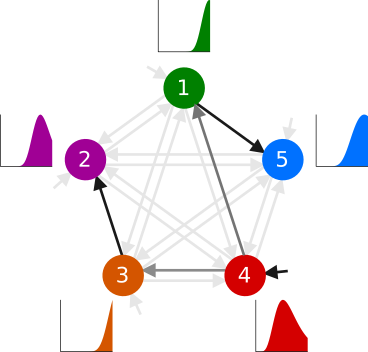}
		\caption{\textbf{Left:} Learned 5-states HSMM in the global frame for skill \textquote{pick} in 2-D, where demonstrations are labeled in color by the associated states. 
			The model has a single initial state (number 4 in the right graph) from which two branches (states 1-5 and states 3-2) encode the different instances of the skill. 
			\textbf{Right:} transition and duration functions of the HSMM. An arrow's color intensity is proportional to the learned transition probability, where black and light gray respectively depict high and low probabilities.}
		\label{fig:hsmm}
		\vspace{-0.4cm}
	\end{figure}

	The approach introduced in Section~\ref{subsec:hsmm} can not be directly applied to solve Problem~\ref{prob:reproduce-given-final} since the forward variable in~\eqref{eq:forward_variable} computes the sequence of \emph{marginally} most probable states, while we are looking for the \emph{jointly} most probable sequence of states given $\vec \xi_1$ and $\vec \xi_{T}$. 
	As a result, when using~\eqref{eq:forward_variable} there is no guarantee that the returned sequence $\vec s^\star_{T}$ matches both the spatio-temporal patterns of the demonstrations and the initial/final observations.
	In terms of the example in Fig.~\ref{fig:hsmm}, it may return the lower branch as the most likely sequence (due to e.g., more demonstrations), i.e., ``pick from the side'', even if the desired final configuration as the final observation is that the end-effector is on the top of object.

	To overcome this issue, we rely on a modification of the Viterbi algorithm~\cite{forney1973viterbi}.
	The classical Viterbi algorithm has been extensively used to find the most likely sequence of states (also called the Viterbi path) in classical HMMs that result in a given stream of observed events.
    Our method differs from it in two main aspects: \textbf{(a)} it works on HSMM instead of HMM; and more importantly \textbf{(b)} most observations are missing except those at the first and the last time instants.
	More specifically, in the absence of intermediate observations the Viterbi algorithm becomes
	\begin{equation} \label{eq:viterbi}
	\begin{aligned}
		\delta_t(j) &= \max_{d\in\mathcal{D}}\max_{i\neq j} \delta_{t-d}(i) a_{ij} p_j(d)  \prod_{t'=t-d+1}^t \tilde{b}_j(\vec \xi_{t'})\,,\\
		\delta_1(j) &= b_j(\vec \xi_1)\pi_jp_j(1),
		%v_T(j) &=&  b_j(\vec o_T)\max_{d\in\mathcal{D}}\max_{i\neq j} v_{T-d}(i) a_{ij} p_j(d) ,\\
	\end{aligned}
        \end{equation}
	where $p_j(d) = \normal(d|\mu_j^D, \sigma_j^D)$ is the duration probability of state $j$, $\delta_t(j)$ is the likelihood of the system being in state $j$ at time $t$ \emph{and} not in state $j$ at $t+1$, see~\cite{forney1973viterbi} for details; and
	\begin{equation*}
	\tilde{b}_j(\vec \xi_{t'}) = \begin{cases}
	\normal(\vec \xi_{t'}|\hat{\vec \mu}_j,\hat{\vec \Sigma}_j), &\, t = 1 \vee t = T;\\
	1, &\, 1 < t < T\,.
        \end{cases}
    \end{equation*}
    where $(\hat{\vec \mu}_j,\hat{\vec \Sigma}_j)$ is the global Gaussian component~$j$ in $\tphsmm_{\mathsf{a}}$ from~\eqref{eq:gaussian-product} given $\vec \xi_{t'}$.
	Namely, at each time $t$ and for each state $j$, the two arguments that maximize equation $\delta_t(j)$ are recorded, and a simple backtracking procedure is used to find the most likely state sequence $\vec s^\star_T$.
	%Note that $T$ is the average duration of the skill.
	In other words, the above algorithm derives the most-likely sequence $\vec s^\star_T$ for skill $\mathsf{a}$ that yields the final observation $\vec \xi_T$, starting from $\vec \xi_1$.

	%%%%%%%%%%%%%%%%%%%%%%%%%%%%%%%
	\subsubsection{\textbf{Trajectory tracking on Manifolds}}\label{subsubsec:reproduce-traj-single}
	Given ${\vec{s}}^\star_T$, we rely on linear quadratic tracking (LQT) to retrieve the optimal reference trajectory.
	We use a linear double integrator dynamics in the control formulation such that the end-effector follows a virtual spring-damper system.
	This allows us to compute an optimal and smooth reference trajectory in closed form without knowledge of the actual robot dynamics.
    As mentioned, the robot state $\vec{x}_{\mathsf{e}}$ lies in the Riemannian manifold ${\mathcal{M}_R = \mathbb{R}^3 \times \mathcal{S}^3 \times \mathbb{R}^1}$, which is prohibitive to define the required linear dynamics.
	However, as in \cite{Zeestraten17riemannian}, we can exploit the linear tangent spaces to achieve a similar result. 
	We assume the linear double-integrator dynamics is defined in the tangent space of $\vec x$, namely $\mathcal{T}_{\vec x}\mathcal{M}_R$.
	Also, we define the tangent space robot state as $\vec{\mathfrak{x}}_t = [\text{Log}_{\vec x_t}(\vec x_t)^\trsp,{\vec v}_t^\trsp]^\trsp\in\mathcal{T}_{\vec x_t}\mathcal{M}_R$, with velocity ${\vec v}_t$ and ${\vec v}_1 = \vec 0$.
	Due to the differential formulation and the definition of the Log-map the robot state in its own tangent space becomes $\vec{\mathfrak{x}}_t = [\vec 0^\trsp, {\vec v}_t^\trsp]^\trsp$.
	Then, the optimal control problem becomes
	\vspace{-.1cm}
	\begin{equation*}
          \begin{aligned}
	& \vec u^\star = \arg\min_{\vec u} \sum_{t=1}^T \left(\text{Log}_{\vec x_t}\!(\vec{\hat{\mu}}_{k_t})^\trsp \vec{\hat{\Sigma}}_{k_t}^{-1}\text{Log}_{\vec x_t}\!(\vec{\hat{\mu}}_{k_t}) \right. \left. + \vec u_t^\trsp \vec R \vec u_t \right), \\
&\mbox{s.t.}\quad \vec{\mathfrak{x}}_{t+1} = \vec A \vec{\mathfrak{x}}_t + \vec B \vec u_t, ~~\vec{\mathfrak{x}}_{t+1}, \vec{\mathfrak{x}}_t, \vec u_t \in  \mathcal{T}_{\vec x_t}\mathcal{M}_R; \\
&\qquad 	\vec x_{t+1} = \text{Exp}_{\vec x_t}(\vec{\mathfrak{x}}_{t+1}) \in \mathcal{M}_R ,~ \vec x_1 = \vec x_{\mathsf{e}}\\
& \qquad  \vec A = \left[ \begin{array}{cc}
\vec I & \vec I \Delta t \\
\vec 0 & \vec I \\
\end{array} \right], ~\vec B = \left[ \begin{array}{c} \vec 0 \\ \vec I \Delta t \end{array} \right]
        \end{aligned}
          \end{equation*}
	where $k_t$ is the $t$-th component in ${\vec{s}}^\star_T$. 
	Specifically, the state error between the desired reference $\vec{\hat{\mu}}_{k_t}$ and current robot state $\vec x_t$ is computed using the logarithmic map $\text{Log}_{\vec x_t}\!(\vec{\hat{\mu}}_{k_t})$ that projects the minimum length path between $\vec{\hat{\mu}}_{k_t}$ and $\vec x_t$ into $\mathcal{T}_{\vec x_t}\!\mathcal{M}_R$. 
	We assume $\vec 0$ reference velocity, therefore we omit this component in the cost funciton.
	The covariance matrices $\vec{\hat{\Sigma}}_{k_t}$ describe the variance and correlation of the robot state variables in a tangent space $\mathcal{T}_{\hat{\vec \mu}_{k_t}}\!\mathcal{M}_R$. 
        Such covariance matrices must be rotated via parallel transport on the manifold to avoid distortion (see Sec.~\ref{sec:riemannian}).
        Similarly, when we propagate the velocity between consecutive states we have to parallel transport them. 
	For the considered manifold $\mathcal{M}_{R}$, this can be computed in closed form~\cite{Zeestraten17riemannian}. 

	%% Let us denote the parallel transport from $\hat{\vec \mu}_{k_t}$ to $\vec x_t$ as $\transp{\hat{\vec \mu}_{k_t}}{\hat{\vec x}_{t}} $, then the transported covariances are $\vec{\tilde{\Sigma}}_{k_t} = \transp{\hat{\vec \mu}_{k_t}}{\hat{\vec x}_{t}} \vec{\hat{\Sigma}}_{k_t} \transp{\hat{\vec \mu}_{k_t}}{\hat{\vec x}_{t}}^\trsp \in \mathcal{T}_{\vec x_t}\mathcal{M}_R.$
	%% Similarly, when the tangent space robot state $\vec v_t$ contains higher order terms, such as velocities, they also have to be transported between consecutive states, e.g., $\tilde{\dot{\vec v_t}} = \transp{\vec x_{t-1}}{\vec x_{t}} \dot{\vec v}_t$, 
	%% with $\tilde{\dot{\vec v_t}} \in \mathcal{T}_{\vec x_t}\mathcal{M}_R$, 
	%% $\dot{\vec v_t} \in \mathcal{T}_{\vec x_{t-1}}\mathcal{M}_R$ and 
	%% $\Vert \tilde{\dot{\vec v_t}}\Vert = \Vert \dot{\vec v_t} \Vert$.
    %For more information we refer to \cite{Zeestraten17riemannian}.
        
        The derivation of the optimal control $\vec u^\star$ and state $\vec x^\star$ trajectory follows the ideas of classical optimal control.
        Similarly to standard LQT, we derive a recursive computation of a feedback and feedforward controller to satisfy the Bellman equation.
        As the control signal and the controller gains are defined in the tangent space of the current robot state, we have to ensure that each variable is parallel transported to this space during recursion.
        %This only requires the computation of the parallel transport between each time step and a few vector-matrix multiplications.
        Overall, the Riemannian extension of LQT requires only a minor computational overhead.

	%========================================

	%========================================
	\subsection{Skill Sequence Reproduction} \label{subsec:sequence-reproduce}
	The above method would suffice for reproducing a \emph{single} skill, which however can not be applied directly to solve Problem~\ref{prob:overall-statement}.
	This is because only the final observation after the whole sequence $\vec{\mathsf{a}}^\star$ is given, while the intermediate observations after each skill are lacking.
	To overcome this, we propose a two-step solution: % as follows:
	\textbf{(1)} cascade the models of each skill within $\vec{\mathsf{a}}^\star$ into one \emph{complete} model $\widehat{\tphsmm}_{\vec{\mathsf{a}}^\star}$;
	\textbf{(2)} find the \emph{complete} state sequence $\widehat{\vec{s}}^\star$ within $\widehat{\tphsmm}_{\vec{\mathsf{a}}^\star}$ to reach the goal state with highest probability.
	
    %========================================
	\subsubsection{\textbf{Cascade multiple HSMMs}}\label{subsubsec:cascade-hsmm}
    We first focus our description on cascading two HSMMs, which can then be applied recursively for longer sequences.
    Consider two TP-HSMMs $\tphsmm_{\mathsf{a}_1}$ and $\tphsmm_{\mathsf{a}_2}$, the algorithm for cascading them into $\widehat{\tphsmm}$ is summarized in Alg.~\ref{alg:cascade-two}.
    The key insight is that the same model $\tphsmm_{\mathsf{a}_2}$ is updated \emph{differently} depending on the final component, or terminal state of $\tphsmm_{\mathsf{a}_1}$ to which $\tphsmm_{\mathsf{a}_2}$ is cascaded to.
    This is because each final component encodes different transformations of the task parameters of $\tphsmm_{\mathsf{a}_1}$ after executing $\mathsf{a}_1$, which in turn results in different ways to update the components in $\tphsmm_{\mathsf{a}_2}$.
    Consequently, the composed model $\widehat{\tphsmm}$ has size $K_1 + K_{1,f}\cdot K_2$, where $K_1$ and $K_2$ are the number of components of $\tphsmm_{\mathsf{a}_1}$ and $\tphsmm_{\mathsf{a}_2}$, respectively, while $K_{1,f}$ is the number of final components in $\tphsmm_{\mathsf{a}_1}$.
    More specifically, Alg.~\ref{alg:cascade-two} consists of two main steps:
	\textbf{(a)} compute the transition probability from \emph{each} final component in $\tphsmm_{\mathsf{a}_1}$ to \emph{each} initial component in $\tphsmm_{\mathsf{a}_2}$;
	\textbf{(b)} modify all components of $\tphsmm_{\mathsf{a}_2}$ for each \emph{final} component in $\tphsmm_{\mathsf{a}_1}$ that $\tphsmm_{\mathsf{a}_2}$ is cascaded to.

	To begin with, we recall the precondition and effect model proposed in our earlier work~\cite{Schwenkel2019Optimizing}.
	In particular, the learned precondition model, denoted by $\boldsymbol{\Gamma}_{1,\mathsf{a}}$, contains TP-GMMs for the initial robot state, i.e., 
	%\begin{equation}\label{eq:precondition}
    	$\boldsymbol{\Gamma}_{1,\mathsf{a}} = \{(\vec{\hat{\mu}}^{(p)}_{1}, \vec{\hat{\Sigma}}^{(p)}_{1}),\, \forall p\in P_{1,\mathsf{a}}\}$,
    %\end{equation}
    where $P_{1,\mathsf{a}}$ is the chosen set of task parameters, derived from the initial system state (e.g., initial pose of objects of interest).
    In addition, we introduce here the final condition model $\boldsymbol{\Gamma}_{T,\mathsf{a}}$, which is learned in a similar way as $\boldsymbol{\Gamma}_{1,\mathsf{a}}$, but for the final robot state, i.e., 
    %\begin{equation}\label{eq:endcondition}
    	$\boldsymbol{\Gamma}_{T,\mathsf{a}} = \{(\vec{\hat{\mu}}^{(p)}_{T}, \vec{\hat{\Sigma}}^{(p)}_{T}),\, \forall p\in P_{T,\mathsf{a}}\}$,
  	%\end{equation}
	where $P_{T,\mathsf{a}}$ is the chosen set of frames, derived from the final system state.
    Simply speaking, $\boldsymbol{\Gamma}_{1,\mathsf{a}}$ models the initial configuration before executing skill~$\mathsf{a}$, while $\boldsymbol{\Gamma}_{T,\mathsf{a}}$ models the final configuration afterwards.
    Furthermore, the learned effect model $\boldsymbol{\Gamma}_{1T,\mathsf{a}}$, contains TP-GMMs for the \emph{predicted} final system state, i.e.,
    %\begin{equation}\label{eq:state-map}
    	$\boldsymbol{\Gamma}_{1T,\mathsf{a}} = \big{\{}\{(\vec{\hat{\mu}}^{(p)}_{1, \mathsf{o}}, \vec{\hat{\Sigma}}^{(p)}_{1, \mathsf{o}}),\, \forall p\in P_{1,\mathsf{a}}\},\, \forall \mathsf{o} \in \mathsf{O}_{\mathsf{a}}\cup \mathsf{e}\big{\}}$,
   	%\end{equation}
    where $P_{1,\mathsf{a}}$ is defined in $\boldsymbol{\Gamma}_{1,\mathsf{a}}$.
    Notice the differences among these three models: the task parameters for $\boldsymbol{\Gamma}_{T,\mathsf{a}}$ are computed from the final system state, while those for $\boldsymbol{\Gamma}_{1,\mathsf{a}}$ and $\boldsymbol{\Gamma}_{1T,\mathsf{a}}$ are extracted from the initial system state.
    Their derivation is omitted here and we refer the readers to~\cite{Schwenkel2019Optimizing}.
    For the sake of notation, we define $\boldsymbol{\Gamma}_{\mathsf{a}}\triangleq\{\boldsymbol{\Gamma}_{1,\mathsf{a}}, \boldsymbol{\Gamma}_{T,\mathsf{a}}, \boldsymbol{\Gamma}_{1T,\mathsf{a}}\}$.

	%========================================
	\setlength{\textfloatsep}{5pt}
	\begin{algorithm}[t]
		\caption{Cascading a pair of TP-HSMMs} \label{alg:cascade-two}
		\LinesNumbered
		\DontPrintSemicolon
		\KwIn{$(\tphsmm_{\mathsf{a}_1}, \boldsymbol{\Gamma}_{\mathsf{a}_1})$ and $(\tphsmm_{\mathsf{a}_2}, \boldsymbol{\Gamma}_{\mathsf{a}_2})$.}
		\KwOut{$(\widehat{\tphsmm}, \widehat{\boldsymbol{\Gamma}})$}
		\ForAll{final component $k_f \in \tphsmm_{\mathsf{a}_1}$}
		{
		Create copy of $\tphsmm_{\mathsf{a}_2}$ as $\tphsmm^{k_f}_{\mathsf{a}_2}$.\;
		Compute $\{a_{k_f, k_i}\}$ for all initial $k_i \in \tphsmm^{k_f}_{\mathsf{a}_2}$  by~\eqref{eq:transition-prob}.\;
		Update $\tphsmm^{k_f}_{\mathsf{a}_2}$ and $\boldsymbol{\Gamma}^{k_f}_{1T, \mathsf{a}_2}$ by~\eqref{eq:update-gmm}.\;
		Cascade $\tphsmm_{\mathsf{a}_1}$ and $\tphsmm^{k_f}_{\mathsf{a}_2}$. Add to $\widehat{\tphsmm}$.\;
		}
		Set additional parameters of $\widehat{\tphsmm}$.\;
		$\widehat{\boldsymbol{\Gamma}} = \{\widehat{\boldsymbol{\Gamma}}_{1}, \widehat{\boldsymbol{\Gamma}}_{T}, \widehat{\boldsymbol{\Gamma}}_{1T}\}=\{\boldsymbol{\Gamma}_{1, \mathsf{a}_1}, \boldsymbol{\Gamma}_{T, \mathsf{a}_2}, \{\boldsymbol{\Gamma}^{k_f}_{1T, \mathsf{a}_2}, \forall k_f\}\}$.\;
		\vspace{-0.1cm}		
	\end{algorithm}
	%========================================

	Then, the transition probability from one final component~$k_f$ of $\tphsmm_{\mathsf{a}_1}$ to one initial component $k_i$ of $\tphsmm_{\mathsf{a}_2}$ is:
	\begin{equation}\label{eq:transition-prob}
		a_{k_f,k_i} \,\propto\, \exp\Big{(}- \sum_{p\in P_c} KL\big{(} \boldsymbol{\Gamma}^{(p)}_{T,\mathsf{a}_1}(k_f)||\boldsymbol{\Gamma}_{1,\mathsf{a}_2}^{(p)}(k_i)\big{)}\Big{)},
		\vspace{-.1cm}
	\end{equation}
	where~$KL(\cdot||\cdot)$ is the KL-divergence from~\cite{hershey2007approximating}, $\boldsymbol{\Gamma}^{(p)}_{T,\mathsf{a}_1}(k_f)$ is the GMM associated with component $k_f$ for frame $p$, $\boldsymbol{\Gamma}^{(p)}_{1,\mathsf{a}_2}(k_i)$ is the GMM associated with component $k_i$ for frame $p$; $P_c=P_{T,\mathsf{a}_1}\cap P_{1,\mathsf{a}_2}$ is the set of \emph{common} frames shared by these two models, which can be forced to be nonempty by always adding the global frame.
	This process is repeated for all pairs of final components in $\tphsmm_{\mathsf{a}_1}$ and initial components in $\tphsmm_{\mathsf{a}_2}$.
	Note that the out-going probability of any final component in $\tphsmm_{\mathsf{a}_1}$ should be normalized.

    Secondly, given one final component $k_f$ of $\tphsmm_{\mathsf{a}_1}$, each component $k$ of $\tphsmm_{\mathsf{a}_2}$ should be affine-transformed as follows:
	\begin{equation}\label{eq:update-gmm}
		(\vec{\hat{\mu}}_{k}^{(\hat{p})},\, \vec{\hat{\Sigma}}_{k}^{(\hat{p})}) \triangleq (\vec\mu_k^{(p)},\,\vec\Sigma_k^{(p)}) \otimes (\vec b_{k_f}^{(\hat{p})},\,\vec A_{k_f}^{(\hat{p})}),
		\vspace{-.1cm}
	\end{equation}
  	where the operation $\otimes$ is defined as the same operation of~\eqref{eq:gaussian-product};
    $(\vec b_{k_f}^{(\hat{p})},\vec A_{k_f}^{(\hat{p})})$ is the task parameter computed from the \emph{mean} of $\boldsymbol{\Gamma}_{1T,\mathsf{a}_1}^{(\hat{p}),\mathsf{o}}(k_f)$, where $\mathsf{o}$ is the object associated with the old frame $p$ in $\tphsmm_{\mathsf{a}_1}$ and $\hat{p}$ is the new frame in $\boldsymbol{\Gamma}_{1T,\mathsf{a}_1}^{\mathsf{o}}(k_f)$.
    Note that the change of frames is essential to compute directly all components of $\tphsmm_{\mathsf{a}_2}$ given an initial system state of $\tphsmm_{\mathsf{a}_1}$.
    The same process is also applied to each component of $\boldsymbol{\Gamma}_{1T,\mathsf{a}_2}$ by changing its frames based on $\boldsymbol{\Gamma}_{1T,\mathsf{a}_1}^{\mathsf{o}}(k_f)$.

  	Lastly, other model parameters of $\widehat{\tphsmm}$ such as duration probabilities, initial and final distributions are set with minor changes from $\tphsmm_{\mathsf{a}_1}$ and $\tphsmm_{\mathsf{a}_2}$.
    For instance, the duration probability of $\tphsmm_{\mathsf{a}_2}$ is duplicated to $k_f$ multiple copies;  
    the initial distributions $\tphsmm_{\mathsf{a}_2}$ are set to zero as the initial states of $\widehat{\tphsmm}$ correspond to those of the first model $\tphsmm_{\mathsf{a}_1}$; the final components of $\tphsmm_{\mathsf{a}_1}$ are removed since the final states of $\widehat{\tphsmm}$ are now those of $\tphsmm_{\mathsf{a}_2}$ updated to its multiple instances.
             
  	Now consider the desired skill sequence given as $\vec{\mathsf{a}}^\star=\mathsf{a}_1\mathsf{a}_2\mathsf{a}_3\cdots \mathsf{a}_N$.
    First, Alg.~\ref{alg:cascade-two} is applied to $(\tphsmm_{\mathsf{a}_1}, \boldsymbol{\Gamma}_{\mathsf{a}_1})$ and $(\tphsmm_{\mathsf{a}_2}, \boldsymbol{\Gamma}_{\mathsf{a}_2})$, yielding $(\widehat{\tphsmm}, \widehat{\boldsymbol{\Gamma}})$.
   	Then, Alg.~\ref{alg:cascade-two} is applied to $(\widehat{\tphsmm}, \widehat{\boldsymbol{\Gamma}})$ and $(\tphsmm_{\mathsf{a}_3}, \Gamma_{\mathsf{a}_3})$.
   	This process repeats itself until $\mathsf{a}_N$ as the end of $\vec{\mathsf{a}}^\star$, yielding $(\widehat{\tphsmm}_{\vec{\mathsf{a}}^\star}, \widehat{\boldsymbol{\Gamma}}_{\vec{\mathsf{a}}^\star})$.
  	Note that $\widehat{\tphsmm}_{\vec{\mathsf{a}}^\star}$ is the \emph{complete} cascaded model in the standard TP-HSMM format.

	\subsubsection{\textbf{State sequence generation and tracking}}\label{subsubsec:sequence-reproduce}
	The derived model $\widehat{\tphsmm}_{\vec{\mathsf{a}}^\star}$ is used to reproduce the skill sequence~$\vec{\mathsf{a}}^\star$ as follows.
 	Firstly, the initial system state $\vec{x}_1$ is obtained (e.g., from perception) and mapped to the initial observation $\vec{\xi}_1$. The goal system state $\vec{x}_{\mathsf{G}}$, e.g., from a high-level planner, is also mapped to the final observation $\vec{\xi}_T$.
	The total length~$T$ is then set to, e.g., the accumulative average length of all skills in~$\vec{\mathsf{a}}^\star$.
	Then, the same solution for Problem~\ref{prob:reproduce-given-final} is used to find the most-likely  sequence $\widehat{\vec{s}}^\star$ within model $\widehat{\tphsmm}_{\vec{\mathsf{a}}^\star}$, given $\vec{\xi}_1$ and $\vec{\xi}_T$.
   	Note that $\widehat{\vec{s}}^\star$ contains the sub-sequence of components, denoted by $\widehat{\vec{s}}_h^\star$, to be followed for \emph{each} skill $\mathsf{a}_h\in \vec{\mathsf{a}}^\star$.
  	During the task execution of each skill $\mathsf{a}_h\in \vec{\mathsf{a}}^\star$, firstly the current system state $\vec{x}_{h,0}$ is observed and used to compute the task parameters and update the global GMMs associated with the components in~$\widehat{\vec{s}}_h^\star$.
   	Afterwards, the trajectory tracking control described in Sec.~\ref{subsubsec:reproduce-traj-single} is used to track~$\widehat{\vec{s}}_h^\star$.
    This process repeats for the subsequent skills until the end of $\vec{\mathsf{a}}^\star$, as summarized in Alg.~\ref{alg:summary}.
    Note that simply tracking $\widehat{\vec{s}}^\star$ without observing the actual intermediate system state $\vec{x}_{h,1}$ would often fail, due to the perception error and motion noise.

%========================================
\begin{algorithm}[t]
\caption{Execute skill sequence~$\vec{\mathsf{a}}^\star$} \label{alg:summary}
\LinesNumbered
\DontPrintSemicolon
\KwIn{$\mathsf{A}$,~$\vec{\mathsf{a}}^\star$,~$\vec{x}_1, \vec{x}_{\mathsf{G}}$, and $\{(\tphsmm_{\mathsf{a}}, \boldsymbol{\Gamma}_{\mathsf{a}}), \forall \mathsf{a} \in \vec{\mathsf{a}}^\star\} $.}
%\For(\tcp*[f]{Offline Learning}){each $\mathsf{a}\in \vec{\mathsf{a}}^\star$}
%{
%Learn models~$\tphsmm_{\mathsf{a}}$ and $\boldsymbol{\Gamma}_{\mathsf{a}}$.\tcp*{see~\cite{Schwenkel2019Optimizing}}
%}
Compute the composed model $\widehat{\tphsmm}_{\vec{\mathsf{a}}^\star}$ using Alg.~\ref{alg:cascade-two}.\;
Compute optimal sequence $\widehat{\vec{s}}^\star$ over $\widehat{\tphsmm}_{\vec{\mathsf{a}}^\star}$, given $\vec{\xi}_1$ and $\vec{\xi}_T$ via~\eqref{eq:viterbi}.\;
\For(\tcp*[f]{On-line Execution}){each $a_h\in \vec{\mathsf{a}}^\star$}
{
Observe the current system state~$\vec{x}_{h,1}$.\;
Update the global components in $\widehat{\vec{s}}_h^\star$ given $\vec{x}_{h,1}$.\;
Track $\widehat{\vec{s}}_h^\star$ by motion control till the end. 
}
\end{algorithm}
%========================================
%
	
	%% %% %%%%%%%%%%%%%%%%%%%%%%%%%%%%%%
	%% %% \begin{figure}[th]
	%% %% \centering
	%% %% \includegraphics[width=0.95\linewidth]{figures/cascade_hsmm.png}
	%% %% \caption{The sequenced HSMM corresponding to the high level pick and place plan. Only the first and last skill model produces observations.}
	%% %% \label{fig:cascade_hsmm}
	%% %% \end{figure}
	%% %% %%%%%%%%%%%%%%%%%%%%%%%%%%%%%%

	%% %========================================
	%% \begin{figure}[t]
	%% 	\centering
	%% 	\includegraphics[height=0.48\linewidth]{figures/robot_lab.jpg}
	%% 	\hfill
	%% 	\includegraphics[height=0.48\linewidth]{figures/vision.png}
	%% 	\caption{The experiment set-up (\textbf{Left}) with the Panda arm, the Astra camera and the objects to interact with; the ArUco marker based  vision system (\textbf{Right}).}
	%% 	\label{fig:lab-vision}
	%% \end{figure}
	%% %========================================

	%========================================
	%========================================
	\section{Experiments} \label{sec:experiments}
	We here describe the experiment setup on a 7-DoF robotic manipulator.
	We consider various assembly tasks consisting of different sequences of skills and show how our framework is used to accomplish such tasks from various initial states.
	The pipeline is implemented in C++, Python and ROS.  

	%========================================
   	\subsection{Workspace Setup and Manipulation Tasks}\label{subsec:exp-setup}
\begin{figure}[t]
		\centering
		\includegraphics[height=0.185\linewidth]{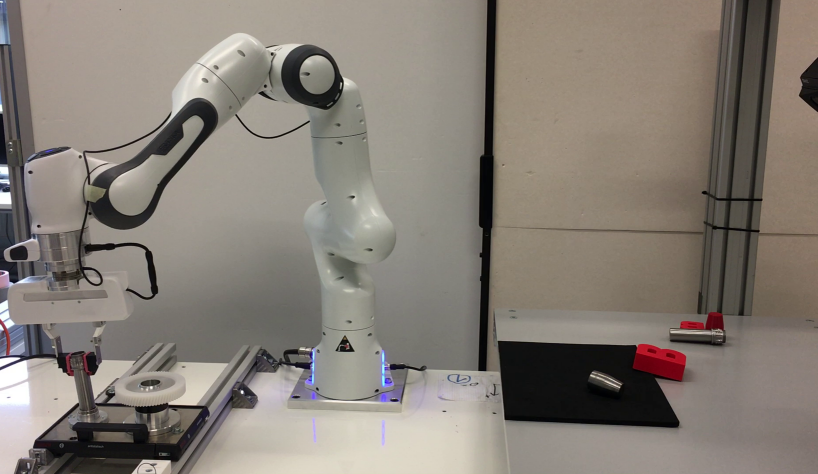}
		\includegraphics[height=0.185\linewidth]{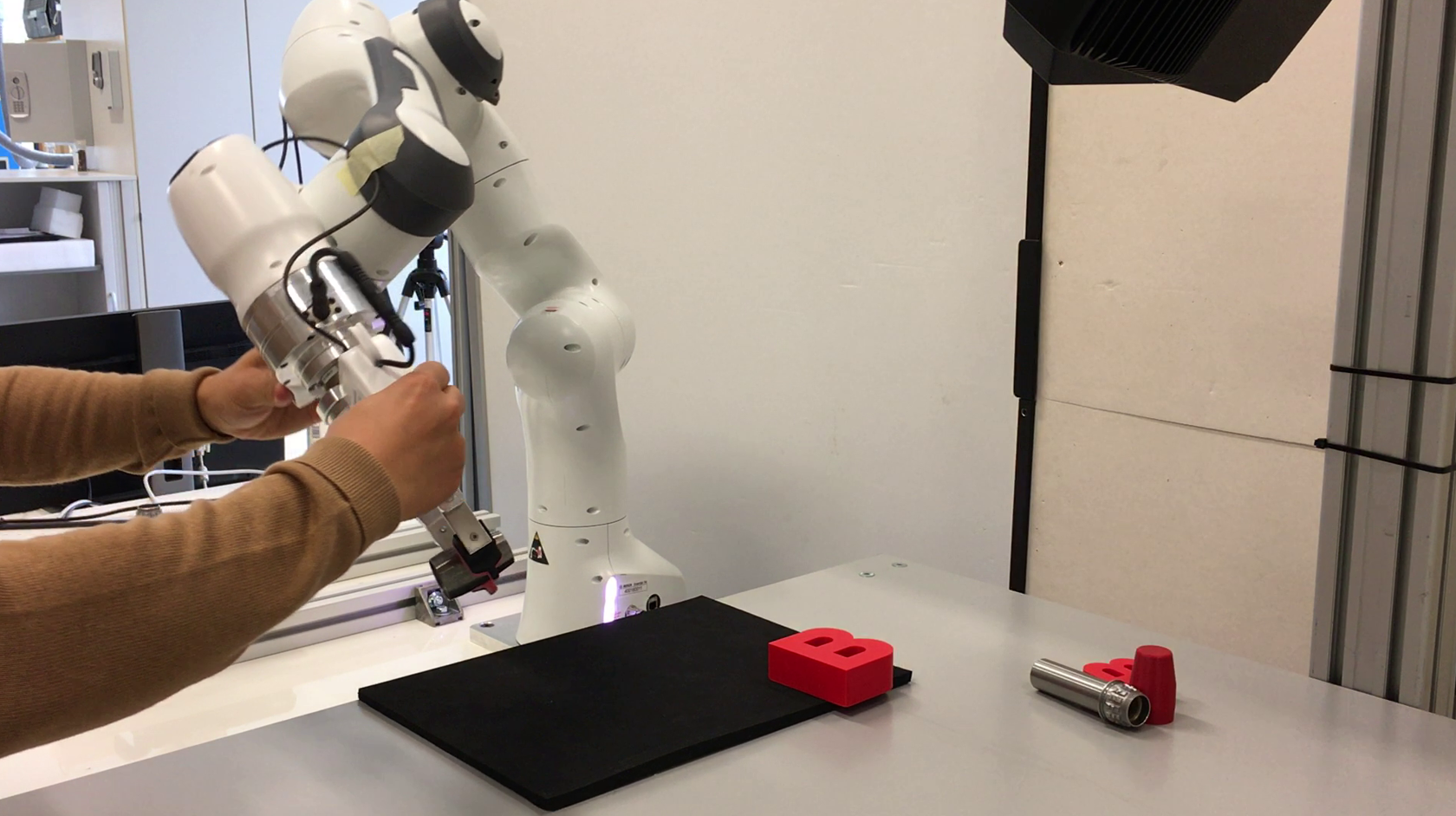}
		\includegraphics[height=0.185\linewidth]{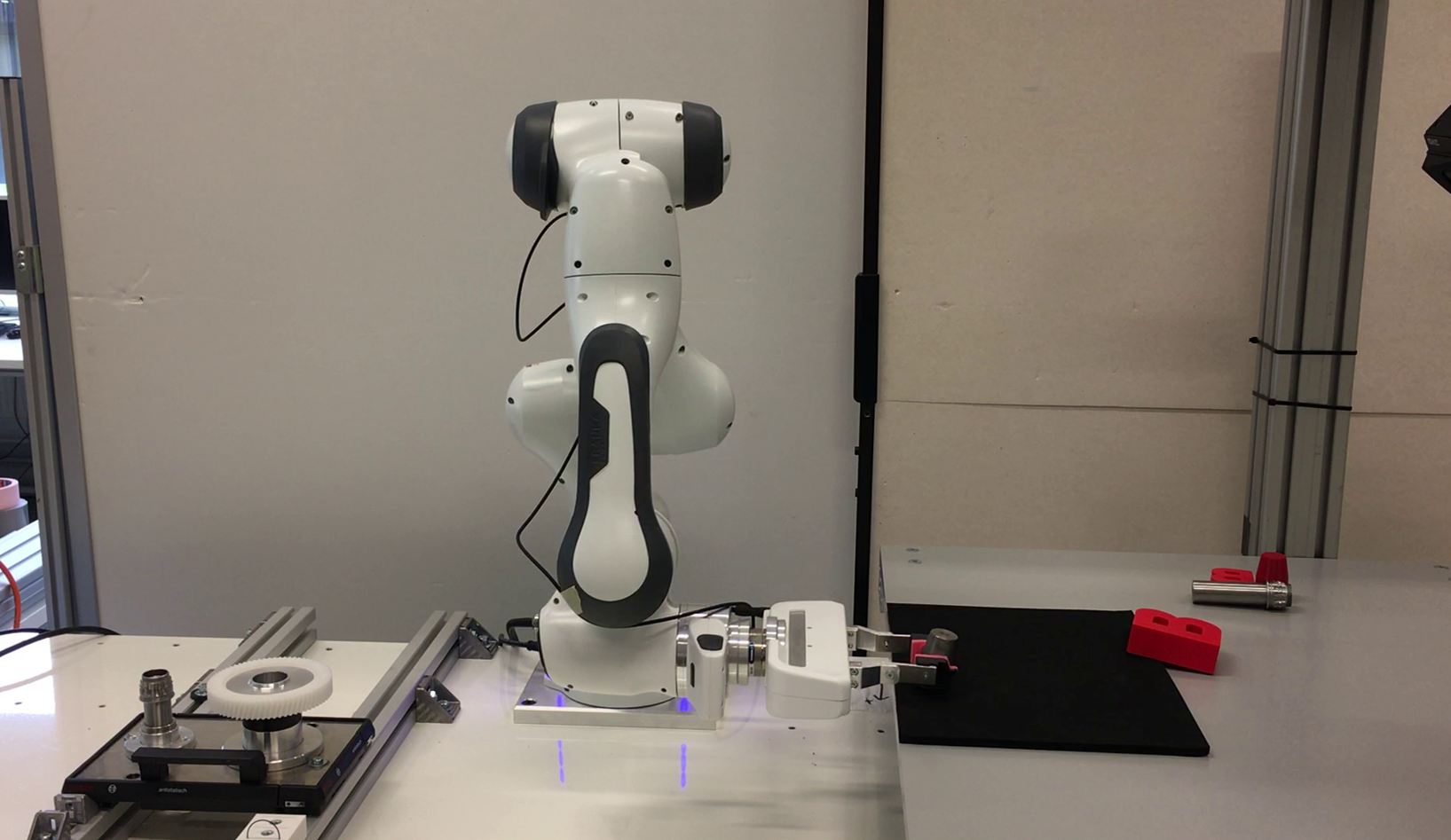}
		%--------------------
		\caption{The experiment setup (\emph{left}), and snapshots of kinesthetic teaching and execution of skills $\mathsf{a}_{\texttt{pc}}~\text{and}~\mathsf{a}_{\texttt{ro}}$ (\emph{middle-right}).}
		\label{fig:snapshots}
		\vspace{-0.25cm}
	\end{figure}

     The Franka Emika Panda robot (denoted by $\mathfrak{r}$) has 7 DoF and is equipped with a two-fingers gripper, as shown in Fig.~\ref{fig:snapshots}.
     The workspace consists of a feeding and inspection platform, where pieces are off-loaded and inspected; and an assembly station where various pieces are assembled into a product.
     The platform is monitored by a Zivid 3D camera, from which the collected point-clouds are inputs to the point-pair-feature detection algorithm~\cite{nixon2019feature}.
     It provides a 6D pose estimation with around \SI{1}{\centi\metre} accuracy w.r.t the global frame (denoted by $\mathfrak{g}$).
     A task-space incomplete impedance controller~\cite{hogan1985impedance} is used to track Cartesian reference trajectories.

	%========================================
	\begin{table}[t]
		%--------------------
		\begin{adjustbox}{height=0.18\linewidth}
			\begin{tabular}{cccccccc}
				\toprule
				$\mathsf{A}$ & Skill Name & $M_{\mathsf{a}}$ & $K_{\mathsf{a}}$  & $B_{\mathsf{a}}$ &  $\mathsf{TP}_{\mathsf{a}}$ & $t\, (\tphsmm_{\mathsf{a}}\,\vert \, \vec{\Gamma}_{\mathsf{a}})$ [\si{\second}]\\ \midrule
				$\mathsf{a}_{\texttt{gr}}$ & \texttt{grasp} & $35$ & $28$ & $4$ &  $\{\mathfrak{c}\}$ & $98\, \vert\,  18 $\\
				$\mathsf{a}_{\texttt{ro}}$ & \texttt{re\_orient} & $18$ & $12$ & $2$ & $\{\mathfrak{c}, \mathfrak{r},\mathfrak{g}\}$  & $54\, \vert\,  12 $\\
				$\mathsf{a}_{\texttt{tl}}$ & \texttt{translate} & $9$ & $8$ & $1$ &  $\{\mathfrak{c}, \mathfrak{r},\mathfrak{g}\}$ & $16\, \vert\,  4 $\\
				$\mathsf{a}_\texttt{at}$ & \texttt{attach} & $7$ & $6$ & $1$ &  $\{\mathfrak{c},\mathfrak{g}\}$ & $12\, \vert\,  4 $ \\
				$\mathsf{a}_\texttt{dp}$ & \texttt{drop} & $31$ & $18$ & $3$ &  $\{\mathfrak{c}, \mathfrak{r},\mathfrak{g}\}$ & $120\, \vert\,  18 $ \\ 
				$\mathsf{a}_\texttt{pk}$ & \texttt{pick} & $6$ & $5$ & $1$ &  $\{\mathfrak{p}, \mathfrak{r}\}$  & $9\, \vert\,  3 $ \\
				$\mathsf{a}_\texttt{is}$ & \texttt{insert} & $5$ & $4$ & $1$ & $\{\mathfrak{p}, \mathfrak{r}\}$  & $10\, \vert\,  5 $ \\
				\bottomrule
			\end{tabular}
		\end{adjustbox}
		%--------------------
		\caption{Demonstrated skills $\mathsf{A}$, number of demonstrations $M_{\mathsf{a}}$, number of components $K_{\mathsf{a}}$, number of branches $B_{\mathsf{a}}$, choice of task parameters $\mathsf{TP}_{\mathsf{a}}$, and training time for $\tphsmm_{\mathsf{a}}$ and $\vec{\Gamma}_{\mathsf{a}}$.}
		\label{table:demo}
	\end{table}
	%========================================        %
	
	We consider parts of an E-bike motor assembly process in our tasks.
	\textbf{Task-1}: pick a non-defective metal cap from the platform, and attach it to the top of a metal peg on the assembly station;
	\textbf{Task-2}: pick a defective metal cap from the platform, and drop it to a container.
	Both the cap~$\mathfrak{c}$ and peg~$\mathfrak{p}$ are components of the e-bike motor.
    During kinesthetic teaching, the state of the end-effector is fetched directly from the on-board control manager.
    Demonstrations are recorded at \SI{50}{\hertz}, while the task-space impedance controller runs at \SI{1}{\kilo\hertz}.
    As summarized in Table.~\ref{table:demo}, we demonstrated in total $7$ skills relevant to the task:
    \texttt{grasp} where the robot grasps $\mathfrak{c}$ in four different cases: via the top, the side, flat-right and flat-left (denoted by $\mathsf{a}_{\texttt{gr}}$);
    \texttt{re-orient} where the robot re-orients $\mathfrak{c}$ from lying flat-left or flat-right to stand-up (by $\mathsf{a}_{\texttt{ro}}$);
   	\texttt{translate} where the robot moves $\mathfrak{c}$ to the edge of the platform (by $\mathsf{a}_{\texttt{tl}}$);
    \texttt{attach} where the robot attaches $\mathfrak{c}$ on the top of $\mathfrak{p}$ (by $\mathsf{a}_{\texttt{at}}$);
    \texttt{drop} where the robot drops $\mathfrak{c}$ into the container facing inwards or outwards (denoted by $\mathsf{a}_{\texttt{dp}}$);
    \texttt{pick} where the robot picks $\mathfrak{p}$ (denoted by $\mathsf{a}_{\texttt{pk}}$);
    \texttt{insert} where the robot inserts $\mathfrak{p}$ (denoted by $\mathsf{a}_{\texttt{is}}$).
   	With these skills, the two tasks above can be re-stated as follows:
    \textbf{Task-1}: If $\mathfrak{c}$ is standing-up, the skill sequence is given by $\vec{\mathsf{a}}_1=\mathsf{a}_{\texttt{pk}}\mathsf{a}_{\texttt{is}}\mathsf{a}_{\texttt{gr}}\mathsf{a}_{\texttt{at}}$; 
    If $\mathfrak{c}$ is lying-flat, the skill sequence is given by $\vec{\mathsf{a}}'_1=\mathsf{a}_{\texttt{pk}}\mathsf{a}_{\texttt{is}}\mathsf{a}_{\texttt{gr}}\mathsf{a}_{\texttt{ro}}\mathsf{a}_{\texttt{gr}}\mathsf{a}_{\texttt{at}}$.
   	\textbf{Task-2}: If $\mathfrak{c}$ is standing-up, the skill sequence is given by $\vec{\mathsf{a}}_2=\mathsf{a}_{\texttt{gr}}\mathsf{a}_{\texttt{tl}}\mathsf{a}_{\texttt{gr}}\mathsf{a}_{\texttt{dp}}$;
   	If $\mathfrak{c}$ is lying-flat, the skill sequence is given by $\vec{\mathsf{a}}'_2=\mathsf{a}_{\texttt{gr}}\mathsf{a}_{\texttt{dp}}$;
	        
    Clearly, the desired skill sequence for different tasks depends on the object state.
    For \textbf{Task-1}, since the skill \texttt{attach} is only taught when the cap is grasped from the top (i.e., not from the side), a cap initially lying flat on the platform needs to be first re-oriented to a stand-up state.
    On the contrary, for \textbf{Task-2}, since the skill \texttt{drop} is only allowed when the cap is grasped from the side (i.e., not from the top), a cap initially standing on the platform needs to be first translated to the edge of the platform, and then grasped from the side.
    The re-orientation and grasping from the side are only taught at the edge of the platform to avoid collisions.
    %Such constraints are very difficult to model explicitly in a standard motion planner.
    %
    %
	
	A video of the \textbf{Task-2} experiments is attached as supplementary material, and a extended version including \textbf{Task-1} results can be found in \href{https://youtu.be/dRGLadt32o4}{https://youtu.be/dRGLadt32o4}.
	
	%========================================
	%========================================

	%========================================
	\subsection{Results}\label{subsec:results}
    Following Alg.~\ref{alg:summary}, both the TP-HSMM model $\tphsmm_{\mathsf{a}}$ and the condition model $\vec{\Gamma}_{\mathsf{a}}$ are learned for each skill described above.
    The total number of components, number of branches, task parameters and training times are listed in Table~\ref{table:demo}.
    Note that skills with several branches have more components and in general take longer to train.
    Thus, such models should be learned off-line before online execution.
    Single skills can be easily reproduced given various initial and final observation by following Sec.~\ref{subsec:skill-reproduce}.
    Table~\ref{table:reproduce-single} summarizes the trajectory length and computation time for the optimal state sequence and optimal control.
    It shows that the modified Viterbi algorithm~\eqref{eq:viterbi} and the optimal tracking control are fast enough to be used for online execution.
    Reproduction of a single skill has almost $100\%$ success rate across various initial configurations that are similar to demonstrated ones.
    For instance, the \texttt{Pick} and \texttt{Insert} skills are retrieved when the cap is placed at different locations on the platform surface.
    In general, when the initial configuration differs drastically from any demonstration, the performance degrades while the resulting trajectory stills resembles similar movement.

    For each skill sequence described for \textbf{Task-1} and \textbf{Task-2}, we follow Sec.~\ref{subsec:sequence-reproduce} to reproduce them under various initial system states.
    Since the sub-sequence $\mathsf{a}_{\texttt{pk}}\mathsf{a}_{\texttt{dp}}$ manipulates only the peg $\mathfrak{p}$ and both skills have only one branch, we treat this sequence independently from other sub-sequences which manipulate only the cap $\mathfrak{c}$.
  	Table~\ref{table:reproduce-sequence} summarizes for each sub-sequence the size of the corresponding composed model $\widehat{\tphsmm}$, the number of its initial and final components, and the time taken to compute $\widehat{\tphsmm}$ and $\vec{s}^\star$, respectively.
	As discussed in Sec.~\ref{subsubsec:cascade-hsmm}, the complexity of $\widehat{\tphsmm}$ grows combinatorially with the number of final states in each skill model.
	For long sequences such as $\vec{\mathsf{a}}'_1$ and $\vec{\mathsf{a}}'_2$, it takes much longer to compute $\widehat{\tphsmm}$ than short sequences such as $\vec{\mathsf{a}}_1$ and $\vec{\mathsf{a}}_2$.
	Note that once such models are computed, they can be saved as a standard TP-HSMM model.
	During execution, the initial system state is obtained, e.g., from the perception system while the goal state is specified directly.
	Then the model $\widehat{\tphsmm}$ is loaded directly without the need for recalling Alg.~\ref{alg:cascade-two}.
	It takes on average $0.5\si{\second}$ to compute the complete sequence $\vec{s}^\star$ given $\widehat{\tphsmm}$ for tasks with two simple skills, and around $17\si{\second}$ for tasks with four complex skills.
	As stated in Alg.~\ref{alg:summary}, each skill $a_h\in \vec{\mathsf{a}}^\star$ is executed by tracking the sub-sequence $\vec{s}_h^\star$ given the current system state.
	Supplementary videos show the reproduction of each task under various system states.

	%========================================
	\begin{table}[t]
	\begin{center}
	%--------------------
		\begin{adjustbox}{width=0.95\linewidth}
		\begin{tabular}{ccc||ccc}
			\toprule
			Skill  & $T$ & $t\, (\vec{s}^\star\,\vert \, \vec{u}^\star)[\si{\second}]$ & Skill  & $T$ & $t\, (\vec{s}^\star\,\vert \, \vec{u}^\star)[\si{\second}]$ \\ \midrule
			$\mathsf{a}_{\texttt{gr}}$ & $105$ & $1.5\, \vert\,  0.2 $ & $\mathsf{a}_{\texttt{ro}}$ & $175$ & $0.6\, \vert\,  0.3 $ \\
			$\mathsf{a}_{\texttt{tl}}$ & $90$ & $0.1\, \vert\,  0.1 $ & $\mathsf{a}_{\texttt{at}}$ & $135$ & $0.5\, \vert\,  0.3 $ \\
			$\mathsf{a}_{\texttt{dp}}$ & $132$ & $0.9\, \vert\,  0.2 $ & $\mathsf{a}_{\texttt{pk}}$ & $116$ & $0.2\, \vert\,  0.2 $ \\
			\bottomrule
		\end{tabular}
		\end{adjustbox}
	%--------------------
	\caption{The trajectory length of $T$ time steps and the average computation time of $\vec{s}^\star$ and $\vec{u}^\star$ for reproducing each single skill.}
	\label{table:reproduce-single}
	\end{center}
	\vspace{-0.35cm}
	\end{table}
	%========================================

	%========================================        %
	\subsection{Discussion}\label{subsec:discussion}
	%There are several discoveries that are worth mentioning during the experiment.
	As shown in Table~\eqref{table:reproduce-sequence}, the computation time grows combinatorially to the number of \emph{final} states for each skill in the sequence.
	So, the more branches there are in one skill, the more time consuming to compose it with other skills.
	This is because all branches are compared in terms of likelihood for both the current skill and the subsequent one.
	Thus our framework, at its present state cannot be used for real-time replanning.
	Nonetheless, several methods can be used to accelerate this process:
    \textbf{(1)} the skill sequence can often be decomposed in independent sub-sequences, which can be planned separately;
    \textbf{(2)} prune early-on transitions whose probability is below a threshold both in the skill model and the transition probability from~\eqref{eq:transition-prob};
    \textbf{(3)} construct $\widehat{\tphsmm}_{\vec{\mathsf{a}}}$ \emph{specifically} for the given initial and final states, i.e., combine~\eqref{eq:transition-prob} with predicted observation probability of $k_i$.
    Also, certain skills may be formulated with \emph{free} task parameters, which can be optimized in a general way for various tasks.
    This issue is not addressed here since all task parameters are assumed to be attached to physical objects.
    Combination of these methods will be addressed in future research.

	%% %========================================
	%% \begin{figure}[t]
	%% 	\centering
	%% 	\includegraphics[width=0.5\linewidth]{figures/illustrativeHSMM_transprobFull1.eps}
	%% 	\caption{The cascaded complete TP-HSMM for a given task, where the transitions from state $1$ to $11$ and from state $4$ to $10$ are created to transit from one skill to another.}
	%% 	\label{fig:complete-hsmm}
	%% \end{figure}
	%% %========================================

	%========================================
	%========================================
	\section{Conclusion} \label{sec:conclusion}
	We presented a framework for learning and sequencing of robot manipulation skills that are object-centric and learned from demonstration.
	In our framework we learn generic skills that encapsulate different instantiations, whose activation rely on the initial and goal state of the system.  
	Given a goal configuration and a manipulation task as skill sequence, our framework computes the optimal state sequence to accomplish the task.
	For successful execution, we also proposed a Riemannian optimal controller in order to calculate a smooth reference trajectory in the manifold of end-effector poses. 
	In contrast to previous works, we experimentally studied our framework on real complex assembly tasks, showing that our approach scales to long sequences of manipulation skills. 
	%\newpage
	%========================================
        	%========================================
	\begin{table}[t]
	\begin{center}
		%--------------------
		\begin{adjustbox}{width=0.95\linewidth}
		\begin{tabular}{cccccc}
			\toprule
			$\vec{\mathsf{a}}^\star$  & $K$ & $E$ & $K_i\,\vert \, K_f$ & $T$ & $t\, (\widehat{\tphsmm}\,\vert \, \vec{s}^\star)[\si{\second}]$ \\ \midrule
			$\mathsf{a}_{\texttt{pk}}\mathsf{a}_{\texttt{is}}$ & $11$& $10$& $1\,\vert\, 1$ & $216$ & $11\, \vert\,  0.1 $ \\
			$\mathsf{a}_{\texttt{gr}}\mathsf{a}_{\texttt{dp}}$ & $100$& $92$& $4\,\vert\, 12$ & $214$ & $21\, \vert\,  1 $ \\
			$\mathsf{a}_{\texttt{gr}}\mathsf{a}_{\texttt{tl}}\mathsf{a}_{\texttt{gr}}\mathsf{a}_{\texttt{dp}}$ & $460$& $440$& $4\,\vert\, 76$ & $435$ & $57\, \vert\,  19 $ \\
			$\mathsf{a}_{\texttt{gr}}\mathsf{a}_{\texttt{at}}$ & $52$& $48$& $4\,\vert\, 4$ & $239$ & $10\, \vert\,  0.4 $\\
			$\mathsf{a}_{\texttt{gr}}\mathsf{a}_{\texttt{ro}}\mathsf{a}_{\texttt{gr}}\mathsf{a}_{\texttt{at}}$ & $492$& $487$& $4\,\vert\, 80$ & $521$ & $76\, \vert\,  17 $\\
			\bottomrule
		\end{tabular}
		\end{adjustbox}
		%--------------------
	\caption{The number of components $K$, edges $E$, initial $K_i$ and final states $K_f$ of the model $\widehat{\tphsmm}_{\vec{\mathsf{a}}}$.
	The trajectory length of $T$ time steps, the total computation time for $\widehat{\tphsmm}_{\vec{\mathsf{a}}}$ and the associated~$\vec{s}^\star$.}
	\label{table:reproduce-sequence}
	\end{center}
	\vspace{-0.35cm}
	\end{table}
	%========================================        %
	%========================================
	\bibliographystyle{IEEEtran}
	\bibliography{references}
	
\end{document}